# Crowd Detection Using Very-Fine-Resolution Satellite Imagery

Tong Xiao, Qunming Wang, Ping Lu, Tenghai Huang, Xiaohua Tong, Peter M. Atkinson

*Abstract*—Accurate crowd detection (CD) is critical for public safety and historical pattern analysis, yet existing methods relying on ground and aerial imagery suffer from limited spatio-temporal coverage. The development of very-fine-resolution (VFR) satellite sensor imagery (e.g., ~0.3 m spatial resolution) provides unprecedented opportunities for large-scale crowd activity analysis, but it has never been considered for this task. To address this gap, we proposed CrowdSat-Net, a novel point-based convolutional neural network, which features two innovative components: Dual-Context Progressive Attention Network (DCPAN) to improve feature representation of individuals by aggregating scene context and local individual characteristics, and High-Frequency Guided Deformable Upsampler (HFGDU) that recovers high-frequency information during upsampling through frequency-domain guided deformable convolutions. To validate the effectiveness of CrowdSat-Net, we developed CrowdSat, the first VFR satellite imagery dataset designed specifically for CD tasks, comprising over 120k manually labeled individuals from multi-source satellite platforms (Beijing-3N, Jilin-1 Gaofen-04A and Google Earth) across China. In the experiments, CrowdSat-Net was compared with five state-of-the-art point-based CD methods (originally designed for ground or aerial imagery) using CrowdSat and achieved the largest F1-score of 66.12% and Precision of 73.23%, surpassing the second-best method by 1.71% and 2.42%, respectively. Moreover, extensive ablation experiments validated the importance of the DCPAN and HFGDU modules. Furthermore, cross-regional evaluation further demonstrated the spatial generalizability of CrowdSat-Net. This research advances CD capability by providing both a newly developed network architecture for CD and a pioneering benchmark dataset to facilitate future CD development. The source code and dataset are available at https://github.com/Tong-777777/CrowdSat-Net.

*Index Terms*—Crowd detection, Very-fine-resolution satellite imagery, Deep learning, Feature enhancement, Feature fusion.

## I. INTRODUCTION

IN recent years, population growth, continued urbanization and rapid economic development in many regions has led to an increase in the frequency of crowd activities in public areas. Such crowds can pose a series of public safety risks, including traffic congestion [1], crowd crushes [2], security incidents [3] and public health risks [4], [5]. Alleviating these risks, for example through crowd control, has become an essential aspect of ensuring public safety in densely populated gatherings. Crowd detection (CD), which involves estimating the location and count of the individuals in a crowd in a specific place [6], [7], [8], plays a crucial role in mitigating these risks. By managing and controlling crowd movements [9], [10] effectively, CD can help reduce the likelihood of accidents and create a safer environment. In addition to crowd management, long-term CD can analyze future crowd distribution patterns by mining historical data trends, providing valuable insights for urban planning and infrastructure optimization [11]. The definition of a crowd varies across research domains. In this research, we adopt a broad definition: A gathering of individuals, whether isolated, dispersed or clustered, or any combination of these, is regarded as a crowd for detection.

To achieve CD effectively, early research relied primarily on analyzing imagery captured by ground surveillance or fixed cameras [12], [13], [14]. These cameras provide diverse perspectives, varying scales and different illumination conditions with which to detect crowds. However, their coverage was inherently limited to specific areas within the camera's field of view. With the development of airborne and unmanned aerial vehicle platforms, aerial datasets, such as the DLR Aerial Crowd Dataset [15], Meynberg's dataset [16] and Mliki's dataset [17], have been introduced for CD, enhancing the development of models to monitor and analyze crowd dynamics over large areas. Although aerial imagery provides many advantages for monitoring, especially its potentially fine spatial resolution, its effectiveness for long-term analysis is constrained by infrequent data updates. For example, in Hebei Province, China, the standard update cycle for aerial photography typically exceeds three years [18], which limits its ability to capture continuous short-term variations in crowd distributions.

With advances in remote sensing satellite technology, fine spatial-temporal resolution imagery has increasingly been applied to detect small-scale (i.e., small area) objects, such as ships, airplanes and tiny buildings [19], [20], [21]. Compared to aerial imagery, satellite sensor imagery (hereafter, satellite imagery) offers a broader spatial coverage and shorter revisit intervals (e.g., the Worldview-4 satellite has a revisit period of approximately 4.5 days [22]), increasing the potential for consistent, large-scale monitoring and historical analysis. Nowadays, very-fine-resolution (VFR) satellite imagery from

This work was supported by the National Natural Science Foundation of China under grant Nos. 42222108, 42221002 and 42171345. *(Corresponding author: Qunming Wang.)*

T. Xiao, Q. Wang, P. Lu, T. Huang and X. Tong are with the College of Surveying and Geo-Informatics, Tongji University, Shanghai 200092, China (e-mail: wqm11111@126.com).

P. Atkinson is with Faculty of Science and Technology, Lancaster University, LA1 4YR Lancaster, U.K., and also with Geography and Environment, University of Southampton, SO17 1BJ Southampton, U.K.

sources like the Beijing-3N (BJ3N) and Jilin-1 Gaofen-04A (JL4A) satellites, as well as the Google Earth platform, has enabled the discrimination of individuals with the naked eye (as shown in Fig. 1). This advance potentially allows more detailed and extensive monitoring, providing a valuable tool for applications requiring accurate object detection and tracking over wide areas. In light of this, as a primary contribution of this research, we presented a novel CD dataset, namely, CrowdSat, derived from the BJ3N and JL4A satellites, as well as the Google Earth platform, covering 32 provincial-level divisions in China (except Guizhou Province and Macao). CrowdSat contains over 120k manually labeled individuals and consists of diverse regions with strong heterogeneity (e.g., built-up areas, snowy regions, beaches, desert regions, etc.). Compared to ground imagery, CrowdSat provides a much broader coverage, making it a more appropriate source for large-scale monitoring. Moreover, with its high revisit frequency, CrowdSat enables more continuous analysis of human activity patterns compared to aerial imagery, which is typically updated less frequently, especially at a large scale. To the best of our knowledge, CrowdSat is the first-ever CD dataset based on VFR satellite imagery, and it is intended to facilitate research on large-scale crowd analysis while uncovering historical patterns of human movement.

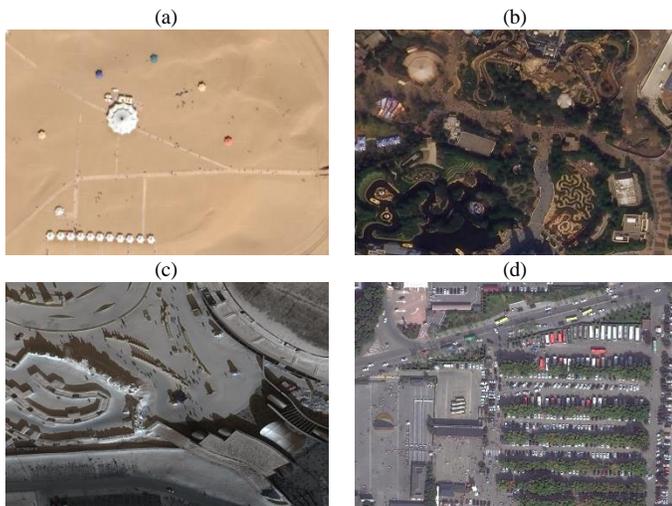

Fig. 1. Examples of VFR satellite imagery for CD. (a) Yuesha Island, Inner Mongolia, China, acquired by the Google Earth platform on Feb. 22, 2023 with a spatial resolution of 0.30 m. (b) Shanghai Disney, China, acquired by the JL4A satellite on Feb. 16, 2023 with a spatial resolution of 0.31 m. (c) Harbin Ice and Snow World on Feb. 2, 2024 and (d) the Xi'an Emperor Qinshihuang's Mausoleum Site Museum in China on Oct. 29, 2023, both acquired by the BJ3N satellite with a spatial resolution of 0.30 m.

Compared to ground and aerial imagery, VFR satellite imagery is more stable and provides with greater consistency through time and can, thus, capture several sudden or unexpected crowd gathering cases. VFR satellite imagery, thus, has a greater range of potential applications, including: 1) Disaster response and emergency management: VFR satellite imagery can be used to monitor the gathering together of people after natural disasters, such as at gathering points for survivors in earthquake areas, temporal shelters in floods, etc..

2) Urban mobility analysis: VFR satellite imagery can be utilized to capture the macro distribution of urban pedestrian flow and reveal the unrecorded flow patterns around transportation hubs, markets and activity places, enabling the identification of temporary stall hotspots to inform improved urban planning. 3) Anomalous crowd gathering detection: VFR technology can detect abnormal population gatherings across a large spatial coverage and analyze the spatio-temporal patterns of their evolution. For example, this capability may be valuable in identifying irregular migrant clusters in sparsely populated border regions. Moreover, it can help track transient poacher activity in wildlife reserves. Furthermore, in urban contexts, spontaneous large gatherings can be identified through temporary infrastructure patterns and anomalous crowd accumulations in non-designated regions. 4) Epidemic prevention and public health surveillance. VFR satellite imagery can help track changes in crowd density in urban centers, transportation hubs and quarantine areas, providing early warning of the potential for disease transmission during disease outbreaks. 5) Large-scale event management. VFR satellite imagery can support (i.e., prepare for) crowd management during large public events, such as sports games, concerts and public rallies. By analyzing historical data, event organizers can optimize pedestrian paths, prevent congestion and improve emergency response measures to ensure safe and smoothly run events. 6) Humanitarian assistance and refugee camp monitoring. VFR satellite-based technology has the potential to contribute to humanitarian assistance efforts by tracking refugee migration patterns, assessing the scale of temporary settlements and assisting international organizations in optimizing the distribution of food, water and medical supplies. 7) Tourism and cultural heritage protection. VFR satellite imagery can help manage visitor flows to cultural heritage sites and tourist destinations, reducing overcrowding and mitigating environmental impacts. These diverse applications highlight the need for advanced CD techniques based on VFR satellite imagery, particularly for tasks requiring precise individual-level information.

In the realm of CD, previous methods focused primarily on estimating crowd counts [23], [24], [25]. However, counting alone provides only basic information, which is insufficient for more comprehensive, higher-level crowd analysis tasks, such as individual tracking, activity recognition, anomaly detection and crowd flow or behavior prediction [26], [27]. To overcome this limitation, recent research has focused increasingly on more challenging fine-grained tasks, such as accurate identification of the localizations of individuals within a crowd [28], [29]. Motivated by this idea, numerous crowd positioning methods have been proposed. These methods use deep learning technologies to segment individual instances [16], generate crowd density maps [30] or use a bounding box to mark the location of each individual [31]. Recently, the idea of treating objects as points has gained popularity in the field of target recognition [32], [33]. This approach has demonstrated reliable accuracy and fast recognition capabilities.

Notably, the aforementioned approaches are employed primarily in images captured by ground surveillance or fixed cameras or aerial vehicles. As illustrated in Fig. 2, individuals in such imagery exhibit typically clear and distinguishable features. However, whether the detection capabilities in the ground and aerial imagery can be transferred to VFR satellite imagery remains challenging. First, the signal of an individual in VFR satellite imagery is weak. As shown in Fig. 1, the signal size of each individual is approximately 3×3 pixels. During the convolution process, particularly in the pooling stages, this small-sized signal can lead to attenuation or even loss of the signal [34], [35], [36]. Second, feature fusion, which integrates the upsampled coarse and fine spatial resolution features, is potentially a useful method to alleviate this issue. However, traditional upsampling methods, such as nearest neighbor and bilinear interpolation, often fail to recover fine spatial details, leading to over-smooth boundaries [37] and misalignment between the high-frequency details in the upsampled coarse spatial resolution features and the fine spatial resolution features [38], making it challenging to detect small-sized objects effectively.

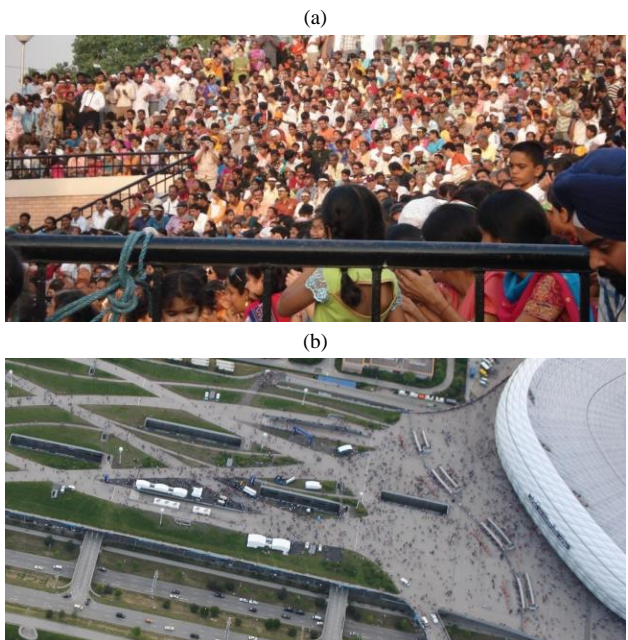

Fig. 2. Examples of ground and aerial imagery. (a) image taken from the ShanghaiTech Dataset[1] and (b) image obtained from the DLR Aerial Crowd Dataset[2].

To overcome the aforementioned limitations, this paper proposed a novel point-based convolutional neural network (CNN) method, CrowdSat-Net, which was specifically designed for large-scale and long-term CD. CrowdSat-Net introduces two key contributions: 1) a Dual-Context Progressive Attention Network (DCPAN) to improve the individual instance feature presentation and 2) a High-Frequency Guided Deformable Upsampler (HFGDU) to replace traditional upsampling methods, which aims to recover the fine spatial information of individuals during the upsampling process.

In summary, the contributions of this research are three-fold：

1) To the best of our knowledge, this is the first research to utilize VFR satellite imagery for CD, which aims to facilitate studies on the characterization of human spatial distributions and temporal activities at a large-scale (both spatially and temporally).
2) To achieve this task, a novel CD dataset, CrowdSat, was collected by multi-source satellite platforms, which comprises over 120k labeled individuals and consists of diverse regions with strong heterogeneity, facilitating the development of CD methods. Additionally, during labeling, a point-like background removal strategy was introduced that uses multi-temporal VFR satellite imagery as auxiliary data to reduce mislabeling rates.
3) A novel point-based CD method using the CNN, termed CrowdSat-Net, was proposed to detect individuals in satellite imagery efficiently. Additionally, two innovational modules, DCPAN and HFGDU, were introduced to enhance individual feature presentation and recover the lost high-frequency information of individual features during upsampling.

The remainder of this paper is structured as follows: Section II provides a comprehensive overview of CrowdSat, detailing its data collection and preprocessing, labeling process and data analysis. Section III presents the architecture of CrowdSat-Net, explaining its key components and design. Section IV presents extensive experimental evaluations, while Section V discusses the broader applicability and inherent limitations of CrowSat and CrowSat-Net. Finally, Section VI summarizes the key findings of this research.

## II. CROWDSAT DATASET

In this paper, CrowdSat was presented for large-scale CD and the analysis of historical human movement patterns. To better understand the details of CrowdSat, this section provides a comprehensive overview, covering three key aspects: data collection and preprocessing, data labeling and data analysis.

### A. Data Collection and Preprocessing

To ensure diverse and representative data for large-scale CD, three complementary remote sensing satellite sources, including the BJ3N and JL04A satellites, and the Google Earth platform, were selected. The BJ3N satellite operates in a Sun-synchronous orbit at an altitude of 610 km, with a 5-day revisit period and an LTAN of 11:00, ensuring consistent and timely coverage. It captures RGB bands with a spatial resolution of 1.20 m and panchromatic bands with a 0.30 m resolution, enabling fine-resolution observation. The JL04A satellite follows a Sun-synchronous orbit at 535 km, with a shorter 3-day revisit period and an LTAN at 10:30, allowing for more frequent observations. The JL04A imagery used in CrowdSat has a panchromatic resolution of 0.31 m and a multispectral resolution of 1.24 m. Additionally, imagery from the Google Earth platform was incorporated with an approximate ground sample distance of 0.3 m.

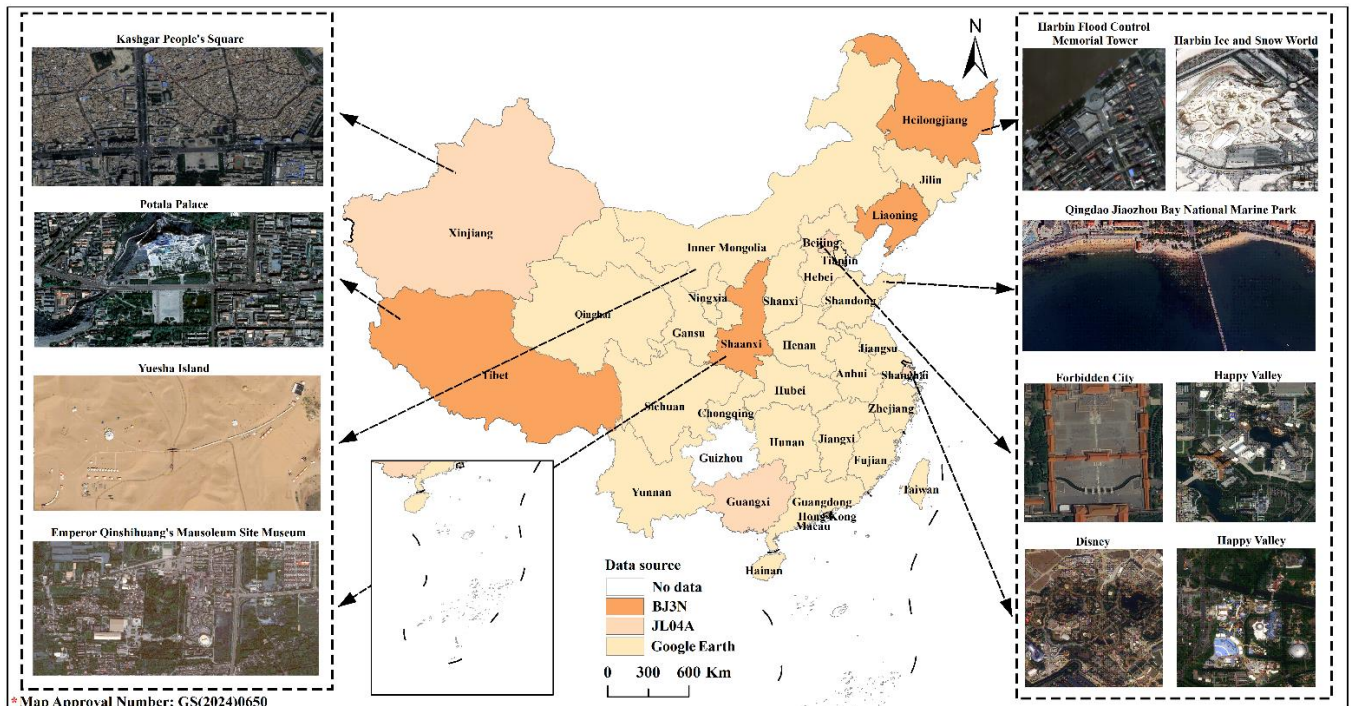

Fig. 3. Spatial distribution map of collected VFR satellite imagery for some locations where crowds typically gather.

To ensure broad spatial coverage and diverse conditions, we selected imagery spanning 32 provincial-level divisions in China (except Guizhou Province and Macao, due to fewer satellite images in these areas). These regions exhibit substantial heterogeneity, encompassing various landscapes and urban settings, as illustrated in Fig. 3. These scenes include open areas (e.g., the Forbidden City), built-up areas (e.g., Kashgar People's Square), snowy regions (e.g., Harbin Ice and Snow World), areas with lush vegetation (e.g., the Emperor Qinshihuang's Mausoleum Site Museum), beaches (e.g., Qingdao Jiaozhou Bay National Marine Park) and desert regions (e.g., Yuesha Island). All the imagery was acquired between Feb. 20, 2023, and Jan. 2, 2025. In addition, before using the collected imagery, preprocessing steps such as geometric correction, clipping and radiometric normalization were performed to enhance the consistency between images from different sources.

Using the original 1.20 m BJ3N or 1.24 m JL4A multispectral imagery directly makes it challenging to distinguish individuals. To address this limitation, a state-of-the-art pan-sharpening technique known as area-to-point regression kriging (ATPRK) [39], [40] was employed to increase the spatial resolution of the BJ3N and JL4A imagery to 0.30 m and 0.31 m, respectively. The fused imagery with finer spatial resolution enables a more accurate interpretation for subsequent analysis.

## B. Data Labeling

Labeling individuals in VFR satellite imagery presents unique challenges compared to ground and aerial imagery. Due to the relatively blurred edges of individuals in VFR satellite imagery, it is difficult to delineate their contours. This increases the likelihood of mistaking other ground objects as individuals. For example, as shown in Fig. 4(a), objects within the red and cyan circles exhibit visual characteristics similar to individuals. However, the objects in the red circles correspond to road asphalt, while those in the cyan circles are street lamps. Similarly, other stationary objects, such as stone pillars, may also be labeled incorrectly, which can further mislead network training.

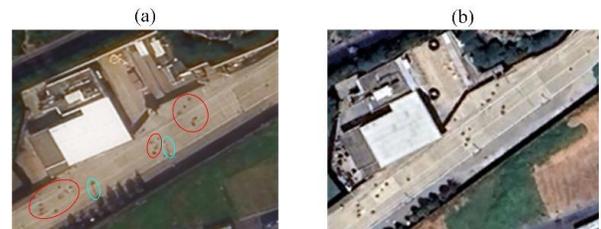

Fig. 4. Examples of mislabeling. (a) and (b) were captured in the same region on Feb. 16, 2023 from the BJ3N satellite and Aug. 7, 2024 from the Google Earth platform, respectively. The objects in the red and cyan circles correspond to road asphalt and street lamps, respectively.

To alleviate the above issue, multi-temporal VFR imagery from the Google Earth platform was used as auxiliary data.

For example, Fig. 4(b) was captured in the same region at a different time, making it easier to distinguish individuals from other small objects. By cross-referencing imagery captured at different time points, fixed objects were identified as non-individuals, reducing labeling errors. Moreover, to further ensure accurate annotations, a center-point labeling strategy was employed. Specifically, for each individual, a single point was placed at the center of the 3×3 pixels region to present the approximate position in the imagery. This approach prevents boundary ambiguity and minimizes the risk of mislabeling non-human objects.

To facilitate the use of the data in a deep learning architecture, the labeled imagery was cropped into 256×256 pixel patches. Patches without individuals were removed, resulting in a total of 3,447 labeled patches.

### C. Data Analysis

Compared to ground and aerial imagery, CrowdSat sets new benchmarks with its vast scale, diverse density distribution and comprehensive multi-environment coverage. It presents four key characteristics.

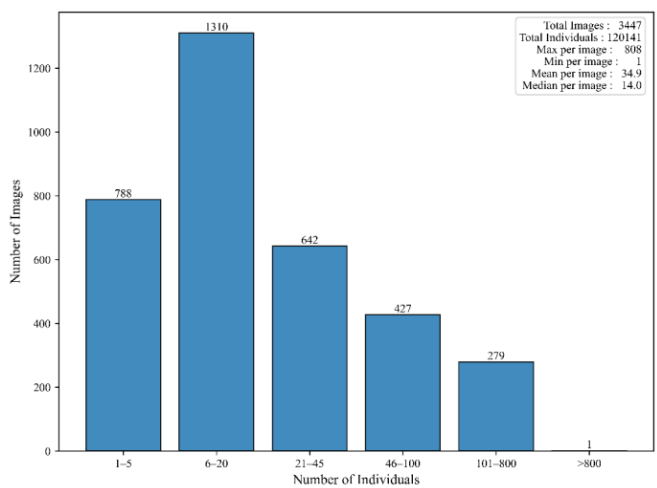

Fig. 5. Crowd count distribution in CrowdSat.

*Extensive National Coverage*. CrowdSat spans China (except Guizhou Province and Macao) and includes samples from 32 Chinese provincial-level divisions. This vast coverage provides a wide range of crowd types, promoting the generalizability of the dataset to various scenarios.

*Multi-Season Adaptation.* CrowdSat was collected between Feb. 20, 2023 and Jan. 2, 2025, covering all four seasons. This wide temporal span allows models to adapt inherently to seasonal variations in terms of lighting conditions (e.g., summer glare and winter haze), clothing styles and crowd movement patterns.

*Diverse Environmental Representations*. CrowdSat contains samples from distinct environmental categories across China, such as open areas, snowy regions, beaches, desert regions, etc.. By systematically encompassing scenarios from historical landmarks to modern transportation hubs, this cross-environment integration can alleviate location-specific bias.

*Multi-Density Representation*. As illustrated in Fig. 5, the CrowdSat dataset consists of a total of 3,447 patches, covering 120,141 individuals. The number of individuals in each patch varies significantly, with the maximum number of 808 individuals and the minimum of 1 individual. On average, each patch contains 34.9 individuals, while the median number per patch is 14.0. This wide range of crowd densities across different patches offers a comprehensive representation of various crowd scenarios, making the dataset valuable for evaluating crowd detection methods under diverse conditions.

## III. METHODS

### A. Overview of CrowdSat-Net

To detect crowds in large-scale VFR satellite imagery effectively, a novel point-based CNN, CrowdSat-Net, was proposed. It was specifically designed to address two key challenges: 1) gradual blurring and loss of small object signals during feature extraction and 2) high-frequency information loss during upsampling operations.

An overview of the CrowdSat-Net architecture is illustrated in Fig. 6. At the beginning, the label image is transformed into a Focal Inverse Distance Transform (FIDT) map [41] using the FIDT method. CrowdSat-Net adopts a two-stage stacked Hourglass Network [42] (a typical network used for point feature extraction), with key modifications to increase the accuracy of CD. First, to improve small-object feature representation, a novel module, that is, DCPAN, is embedded in the image preprocessing stage. Second, during the upsampling operation, the traditional method (e.g., bilinear interpolation) is replaced with a novel module, HFGDU, which aims to restore lost high-frequency information. Within each Hourglass Network, a location map is generated, and the loss is computed using the FIDT map and the location map through the Focal Loss function [43], a strategy known as Intermediate Supervision. The total loss $L$ is computed as the sum of Focal Losses from each Hourglass Network. Finally, the location map from the last Hourglass Network is processed using a Local-Maxima-Detection-Strategy (LMDS) to obtain the final individual localizations.

The remainder of this section introduces important components in detail, including the FIDT map, DCPAN, HFGDU, LMDS and model evaluation.

### B. Focal Inverse Distance Transform (FIDT) Map

It is critical to identify individual localization in CrowdSat-Net. Traditional methods, such as binary-like maps [29], segmentation-like maps [44], topological maps [45] and independent instance maps [46], struggle to distinguish overlapping objects in dense crowds due to their reliance on fixed thresholds or semantic boundaries. To overcome this problem, the FIDT [41] method was employed, which enables overlap-free head localization by assigning higher pixel responses closer to head centers, thereby ensuring accurate localization in dense crowds. The FIDT is defined as follows:

$$I = \frac{1}{d(x,y)^{(\alpha \times P(x,y)+\beta)} + C} \qquad (1)$$

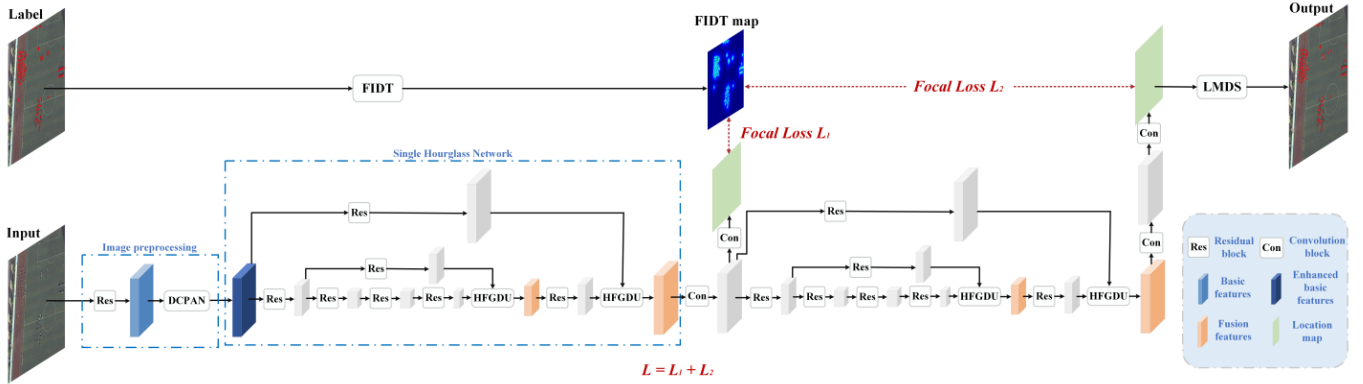

Fig. 6. Overview of the proposed CrowdSat-Net. First, the labeled image is transformed into the FIDT map. During each training iteration, CrowdSat-Net enhances the basic features in the image preprocessing using the DCPAN module. Then, these enhanced features pass through the two-stacked Hourglass Network. In each Hourglass Network, the traditional upsampling method is replaced with the HFGDU module to recover the lost fine details. Each Hourglass Network generates a location map, which is compared with the FIDT map to calculate the Focal Loss. The total loss $L$ is the sum of Focal Losses from each Hourglass Network. After training, the location map conducted by the last Hourglass Network is transformed into the actual localization result using the LMDS method.

where $I$ represents the FIDT map, $\alpha$ and $\beta$ are weight factors, set to the default values of 0.02 and 0.75, respectively [41]. $C$ indicates a constant, which aims to prevent division by zero. $d(x,y)$ denotes the distance between the pixel at location $(x,y)$ and its nearest annotated pixel at location $(x',y')$, in the set of all annotation $\mathbf{B}$:

$$d(x,y) = \min_{(x',y') \in \mathbf{B}} \sqrt{(x-x')^2 + (y-y')^2} \qquad (2)$$

### C. Dual-Context Progressive Attention Network (DCPAN)

The image preprocessing stage aims to extract fundamental feature representations from the original inputs, serving as the foundation for downstream network architectures. Conventional implementations in the Hourglass Network employ typically a standard convolutional layer followed by a ResNet layer [32], which ignores the gradual signal attenuation of small objects. This limitation can compromise model sensitivity to fine-grained features and increase false negative rates in individual detection.

To address the above issue, we first excluded the standard convolutional layer and then proposed the DCPAN module embedded after the ResNet layer, shown in Fig. 7. The DCPAN module combines synergistically a base spatial attention (SA) Encoding and two parallel branches (multi-scale feature extraction (MSFE) and local contrast enhancement (LCE) branches), aiming to improve the small-object signal presentation. Specifically, given a feature map $\mathbf{X} \in \mathbf{R}^{C \times H \times W}$ generated by the ResNet layer, where $C$, $H$ and $W$ denote the channel, height and weight of the feature, respectively, the DCPAN module generates small-object feature enhancement weights through three key operations detailed as follows.

*SA Encoding.* Small objects lack distinct semantic features, but exhibit unique spatial patterns. We first, thus, extracted the

base spatial attention cues $\mathbf{F}_{base}$, achieved as follows:

$$\mathbf{F}_{max} = \max(\mathbf{X}), \ \mathbf{F}_{avg} = \frac{1}{C}\sum_{c=1}^{C} \mathbf{X}_c \qquad (3)$$

where $\mathbf{F}_{max} \in \mathbf{R}^{1 \times H \times W}$ and $\mathbf{F}_{avg} \in \mathbf{R}^{1 \times H \times W}$ represent the max-pooled and average-pooled features, respectively. These two features are then concatenated to form the base spatial feature $\mathbf{F}_{base} \in \mathbf{R}^{2 \times H \times W}$.

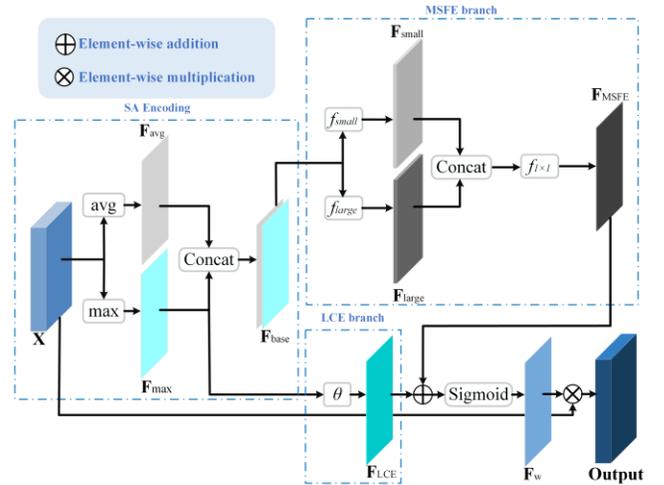

Fig. 7. Flowchart of the DCPAN module, which consists of three main components: SA Encoding, MSFE and LCE branches. The SA Encoding component extracts the base SA feature from the original features $\mathbf{X}$. The MSFE branch captures contextual information from this base feature, while the LCE branch focuses on extracting its local contrast. Finally, the outputs of the MSFE and LCE branches are fused through element-wise addition, and the result is passed through a sigmoid activation function to obtain the enhancement weight map $\mathbf{F}_w$. This map is subsequently applied to the original feature $\mathbf{X}$ via element-wise multiplication.

*MSFE Branch.* Although the base spatial feature can improve the presentation of prominent objects by highlighting global significant regions, the global pooling operations (e.g., max/avg) in SA Encoding fail to model positional dependencies between objects, leading to suboptimal results in cluttered scenes [47]. To alleviate this issue, a simple MSFE branch, which employs parallel dilated convolutional layers with complementary dilation rates, was proposed:

$$\mathbf{F}_{\text{small}} = f_{\text{small}}(\mathbf{F}_{\text{base}}), \mathbf{F}_{\text{large}} = f_{\text{large}}(\mathbf{F}_{\text{base}}) \quad (4)$$

where $f_{\text{small}}$ and $f_{\text{large}}$ are the dilated convolutional operations with dilation rates of 2 and 4, respectively. Then, the $\mathbf{F}_{\text{MSFE}} \in \mathbf{R}^{1 \times H \times W}$ of these layers are concatenated and fused using a $1 \times 1$ convolutional layer:

$$\mathbf{F}_{\text{MSFE}} = f_{1 \times 1}\big(\text{Concat}(\mathbf{F}_{\text{small}}, \mathbf{F}_{\text{large}})\big) \quad (5)$$

*LCE Branch.* Standard pooling operations always blur high-frequency details, leading to degraded accuracy on fine-grained detection tasks [47]. To address this problem, the LCE branch highlights locations with large heterogeneity by a local contrast generator $\theta$:

$$\mathbf{F}_{\text{LCE}} = \theta(\mathbf{F}_{\text{max}}) = \partial(|\mathbf{F}_{\text{max}} - \text{AvgPool}_{3 \times 3}(\mathbf{F}_{\text{max}})|) \quad (6)$$

where $\mathbf{F}_{\text{LCE}} \in \mathbf{R}^{1 \times H \times W}$ represents the contrast-sensitive weight map and $|\cdot|$ denotes absolute value computation. $\partial$ is the learnable contrast enhancement operator composed of a $3 \times 3$ convolutional layer, batch normalization, ReLU activation and a $3 \times 3$ convolutional layer.

Finally, the enhancement weight map $\mathbf{F}_w$ is generated by applying a sigmoid activation to the element-wise addition of $\mathbf{F}_{\text{MSFE}}$ and $\mathbf{F}_{\text{LCE}}$. This weight map is then multiplied pixel-wise with the original feature $\mathbf{X}$ to obtain the final feature.

### D. High-Frequency Guided Deformable Upsampler (HFGDU)

Within each Hourglass network, the coarse spatial resolution features are upsampled and then fused with fine spatial resolution features to increase the ability to detect small objects. However, as mentioned in the Introduction, traditional upsampling methods, such as nearest neighbor and bilinear interpolation, often fail to recover fine spatial details, leading to over-smooth boundaries [37] and misalignment of high-frequency features [38], making it challenging to detect small objects effectively.

To alleviate the above challenge, the HFGDU module was introduced. As shown in Fig. 8, HFGDU alleviates the dual challenges of high-frequency detail recovery and geometric-aware feature alignment through a three-stage architecture: Initial Upsampling, High-Frequency Detail Compensation (HFDC) and Deformable Alignment Fusion (DAF).

*Initial Upsampling.* Given an input coarse spatial resolution feature map $\mathbf{F}_{\text{coarse}} \in \mathbf{R}^{C' \times H' \times W'}$ and a fine spatial resolution feature map $\mathbf{F}_{\text{fine}} \in \mathbf{R}^{C' \times 2H' \times 2W'}$, to match the spatial size, we first applied the simple bilinear interpolation to $\mathbf{F}_{\text{coarse}}$ to obtain an upsampled feature map $\tilde{\mathbf{F}}_{\text{up}} \in \mathbf{R}^{C' \times 2H' \times 2W'}$.

However, bilinear interpolation always smooths feature boundaries, failing to recover the lost high-frequency details that are critical for small object detection.

*HFDC.* To recover the lost fine details during upsampling, we introduced a learnable high-pass filter $f_{\text{hp}}$ to extract high-frequency components from the upsampled feature map:

$$\mathbf{F}_{\text{hf}} = f_{\text{hp}}(\tilde{\mathbf{F}}_{\text{up}}) \quad (7)$$

where the filter $f_{\text{hp}}$ is designed based on a Laplacian-like kernel, which is initialized with a center weight of 1 and surrounding weights of -1/8, to enhance edge structures. The extracted high-frequency feature map $\mathbf{F}_{\text{hf}}$ is then refined using a residual compensation generator $G_{\text{res}}$:

$$\mathbf{F}_{\text{comp}} = G_{\text{res}}(\mathbf{F}_{\text{hf}}) \quad (8)$$

where $G_{\text{res}}$ is composed of a $3 \times 3$ convolutional layer followed by two additional $3 \times 3$ convolutional layers with ReLU activations, and a final $3 \times 3$ convolutional layer with a Sigmoid activation to constrain the compensation magnitude. The map $\mathbf{F}_{\text{comp}}$ is then integrated with the upsampled feature map to generate the compensated high-frequency details map $\mathbf{F}_{\text{HFDC}}$:

$$\mathbf{F}_{\text{HFDC}} = \tilde{\mathbf{F}}_{\text{up}} \otimes (\mathbf{1} + \mathbf{F}_{\text{comp}}) \quad (9)$$

where $\otimes$ means element-wise multiplication. This step helps in recovering the fine high-frequency components, which can potentially increase the clarity and sharpness of the upsampled feature representation.

*DAF.* After compensating for high-frequency details, we employed DAF to refine the spatial alignment between the upsampled feature $\mathbf{F}_{\text{HFDC}}$ and the fine spatial resolution feature $\mathbf{F}_{\text{fine}}$. Traditional upsampling methods rely on fixed-grid sampling, which may cause spatial misalignment. To alleviate this, we introduced a deformable convolution layer $\mathcal{D}$ that predicts spatial offsets $\Delta \mathbf{p}$:

$$\mathbf{F}_{\text{aligned}} = \mathcal{D}(\mathbf{F}_{\text{HFDC}}, \Delta \mathbf{p}) \quad (10)$$

where $\mathbf{F}_{\text{aligned}}$ denotes the deformably sampled feature map used to correct misalignment, and the offsets $\Delta \mathbf{p}$ are adaptively learned from the feature map:

$$\Delta \mathbf{p} = \mathcal{G}(\text{Concat}(\mathbf{F}_{\text{HFDC}}, \mathbf{F}_{\text{fine}})) \quad (11)$$

In Eq. (11), $\mathcal{G}$ is a lightweight offset prediction network composed of two stacked $3 \times 3$ convolutional layers that estimates local displacements. This helps the model to align features dynamically based on the underlying structure.

Furthermore, to integrate selectively aligned features while suppressing redundant information, we introduced a feature modulation gate $\mathbf{M}$ that acts as an adaptive attention mechanism:

$$\mathbf{F_{DAF}} = \mathbf{M} \otimes \mathbf{F_{aligned}} + \mathbf{F_{fine}} \qquad (12)$$

where $\mathbf{F_{DAF}}$ is the final fusion map, and the modulation gate $\mathbf{M}$ is computed as:

$$\mathbf{M} = \sigma(f_{1\times1}(\text{Concat}(\mathbf{F_{HFDC}}, \mathbf{F_{fine}}))) \qquad (13)$$

where $\sigma$ is the sigmoid activation. By integrating these three components, HFGDU aims to preserve high-frequency details and dynamically align features effectively, which can contribute to improving the detection of small objects in the Hourglass Network.

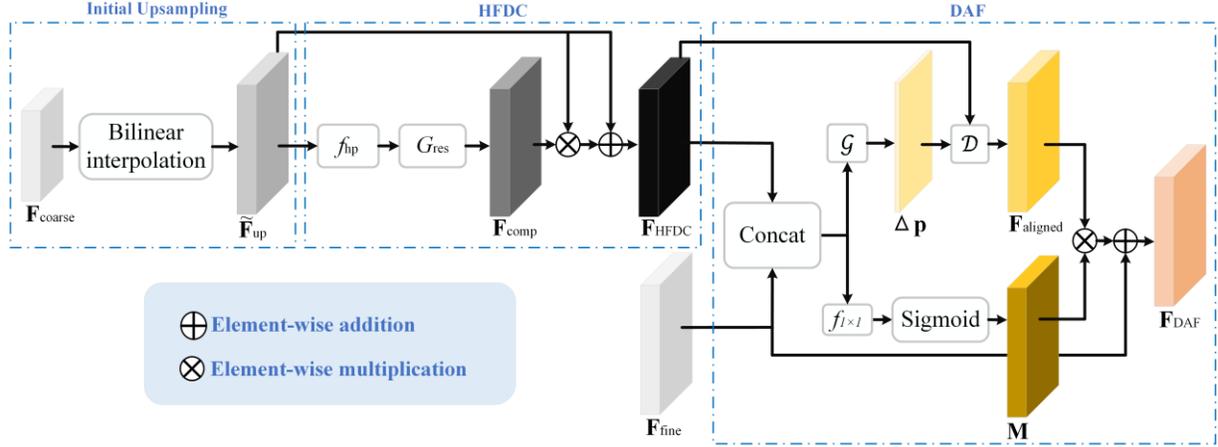

Fig. 8. The structure of the HFGDU module. It contains three main components: Initial Upsampling, HFDC and DAF. Initial Upsampling uses the bilinear interpolation to match the spatial resolution of coarse and fine features. Then, HFDC recovers the high-frequency details of the upsampled features, while DAF refines the spatial alignment of the high-frequency features.

### E. Local Maxima Detection Strategy

The FIDT map generated by the last Hourglass Network can identify the center point of each individual effectively, but it struggles with filtering false positives. To alleviate this issue, the LMDS was introduced to determine the correct position of each individual [41]. Specifically, a 3×3 max-pooling operation is employed to generate all candidate points, and an adaptive threshold, denoted as $\delta$, is used to filter out false positives. This threshold is defined empirically as 100/255.0 times the maximum value of the FIDT map, and only points with values greater than or equal to $\delta$ are selected. Additionally, if the maximum value of the FIDT map is smaller than a tiny fixed threshold (set to 0.10) [41], it indicates that there are no detected individuals in the input image.

### F. Model Evaluation

To analyze the performance of the proposed VRCD model comprehensively, localization metrics are introduced in this research, as they provide fundamental evaluation information. The first step in assessing localization performance is to match the predicted individual point $\hat{P}$ with the ground reference individual point $P$. If the Euclidean distance between $\hat{P}$ and $P$ is less than a distance threshold $\gamma$, they are considered matched. In this study, $\gamma$ was set to 1 pixel [48], and the nearest neighbors method [49] was employed for matching. After matching, the match matrix and counts of the number of True Positive (TP), False Positive (FP) and False Negative (FN) are obtained. Then, Precision, Recall and F1-score [50] are adopted based on TP, FP and FN to evaluate the performance of the proposed model. The formulae are as follows:

$$\text{Recall} = \frac{\text{TP}}{\text{TP+FN}} \qquad (14)$$

$$\text{Precision} = \frac{\text{TP}}{\text{TP+FP}} \qquad (15)$$

$$\text{F1} - \text{score} = \frac{2 \times \text{Recall} \times \text{Precison}}{\text{Recall+Precision}} \qquad (16)$$

### III. EXPERIMENTS

In this section, comprehensive experiments on the CrowdSat dataset were conducted to examine the effectiveness of the proposed CrowdSat-Net method. Specifically, Section III-A details the experimental setup. Section III-B presents a series of ablation studies to validate the contributions of the DCPAN and HFGDU modules. Subsequently, Section III-C compares the CrowdSat-Net method with five recent state-of-the-art CD benchmarks (designed for ground or aerial imagery) to assess its localization performance. Section III-D analyzes the performance of CrowdSat-Net across different crowd density levels. Finally, Section III-E examines the cross-regional generalization of the CrowdSat-Net method by applying it in unseen foreign regions.

### A. Experimental Setup

In this research, the labeled patches in CrowdSat were divided randomly into training and validation sets in a 4:1 ratio. To increase the model's robustness and generalization, we applied augmentation techniques, including horizontal flipping, vertical flipping and CutMix [51]. After augmentation, patches without individuals were removed, resulting in 10,357 training patches and 704 validation patches.

After preparing the dataset, we trained the CrowdSat-Net model and other CD models for 150 epochs using the Adam optimizer [52] with a base learning rate of 0.0003 and a weight decay of 0.001. For all models, the batch size was set to 8, and they were implemented in PyTorch 1.21. All experiments were conducted on a Windows 10 workstation equipped with a 13th Gen Core i7-13700 CPU and a 24 GB NVIDIA GeForce RTX 4090 GPU.

### B. Ablation Experiments

*Ablation of DCPAN and HFGDU.* To examine the contributions of the DCPAN and HFGDU modules in CrowdSat-Net, a basic ablation experiment was conducted. In this experiment, the baseline refers to CrowdSat-Net with the DCPAN module removed and the HFGDU module replaced by bilinear interpolation.

As presented in Table I, the obvious increases in localization metrics are observed with the addition of each module. Specifically, the DCPAN module enhances the feature presentation of small objects in the basic feature extraction process, therefore, leading to an increase in F1-score of 0.90% and in Precision of 4.36% compared to the baseline. Meanwhile, replacing the traditional bilinear interpolation with the HFGDU module increased the F1-score by 1.10% and Precision by 2.09% via its high-frequency information-guided learnable upsampling mechanism. Notably, the integration of both modules achieved the largest F1-score and Precision with F1-score gain of 1.70% and Precision increase of 4.69% compared to the baseline, demonstrating the effectiveness of these two modules.

#### TABLE I
#### BASIC ABLATION STUDY FOR DCPAN AND HFGDU

| Methods | DCPAN | HFGDU | F1-score (%) | Recall (%) | Precision (%) |
|---|---|---|---|---|---|
| Baseline | × | × | 64.42 | 60.77 | 68.54 |
| CrowdSat-Net | √ | × | 65.32 | 59.16 | 72.90 |
| | × | √ | 65.52 | **61.10** | 70.63 |
| | √ | √ | **66.12** | 60.27 | **73.23** |

The best performance is in bold.

*Ablation Study of Different Feature Enhancement Modules in CrowdSat-Net.* To further examine the effectiveness of DCPAN, a comparison with two well-known feature enhancement modules (spatial attention (SA) [53] and efficient channel attention (ECA) [54]) was performed. In this experiment, all components were the same as the baseline method except the feature enhancement module. The result is displayed in Table II. It is seen that DCPAN achieves greater performance than SA and ECA, achieving the largest F1-score of 65.32% (+0.46% over SA and +0.58% over ECA) and Recall of 59.16 (+1.32% over SA and +1.02% over ECA).

Architecturally, DCPAN extends SA by integrating two novel branches: the MSFE and LCE. Specifically, MSFE

utilizes dilated convolutions for multi-scale contextual information, mitigating false negatives effectively caused by the fixed receptive field of SA. Furthermore, in contrast to SA, the LCE enhances feature representation by amplifying local feature responses through contrast-based spatial attention. These innovations increase the localization accuracy of DCPAN collectively. Compared to ECA, which focuses on channel-wise attention exclusively, DCPAN alleviates spatial signal blurring through its dual-branch design.

#### TABLE II
#### COMPARISON OF DCPAN WITH SA AND ECA IN CROWDSAT-NET

| Methods | F1-score (%) | Recall (%) | Precision (%) |
|---|---|---|---|
| SA [53] | 64.86 | 57.84 | **73.82** |
| ECA [54] | 64.74 | 58.14 | 71.65 |
| DCPAN | **65.32** | **59.16** | 72.90 |

The best performance is in bold.

*Ablation of different upsamplers in CrowdSat-Net.* To reveal comprehensively the effectiveness of HFGDU, a comparative analysis with four established benchmark upsamplers - bilinear interpolation, Content-Aware ReAssembly of FEatures (CARAFE) [55], Deconvolution [56] and Dysample [57] - was conducted.

The result in Table III demonstrates that HFGDU exhibits greater localization performance over existing upsamplers, achieving the largest F1-score and Precision. This is attributed to its dual-domain enhancement mechanism: 1) The HFDC module employs a learnable Laplacian operator to amplify boundary gradients, mitigating the inherent edge blurring in bilinear interpolation effectively and yielding a 0.33% Precision gain, a 1.11% Recall increase and a 0.80% F1-score increase over bilinear interpolation. 2) The DAF mechanism based on dynamic migration prediction reduces the feature misalignment error by jointly optimizing the spatial correspondence between coarse and fine spatial resolution features. While Dysample achieves the largest Recall, its Precision lags obviously behind HFGDU (-2.79%). While Deconvolution achieves larger Recall, its fixed transposed convolution kernels lead to checkerboard artifacts, decreasing Precision by 2.44%. Although CARAFE generates upsampled cores via content-sensing, it struggles with small objects in dense scenes, leading to over-smooth results and a lower F1-score (65.17%) than bilinear interpolation (65.32%).

In summary, the above results show that methods that rely solely on dynamic kernels or fixed interpolation kernels will produce a trade-off between Recall and Precision. In contrast, HFGDU achieves simultaneous optimization of both metrics through the co-design of frequency-domain compensation and spatial alignment, making it particularly valuable for dense scene analysis.

#### TABLE III
#### COMPARISON BETWEEN DIFFERENT UPSAMPLERS IN CROWDSAT-NET

| Methods | F1-score (%) | Recall (%) | Precision (%) |
|---|---|---|---|
| Bilinear interpolation | 65.32 | 59.16 | 72.90 |
| CARAFE [55] | 65.17 | 59.47 | 72.08 |
| Deconvolution [56] | 65.38 | 60.74 | 70.79 |
| Dysample [57] | 65.62 | **61.42** | 70.44 |
| HFGDU | **66.12** | 60.27 | **73.23** |

The best performance is in bold.

## C. Comparison Between CrowdSat-Net and Benchmark Methods

This section aims to compare the performance of the proposed CrowdSat-Net with five state-of-the-art CD methods (designed based on ground or aerial imagery): Point to Point Network (P2PNet) [27], SCALNet [58], Point quEry Transformer (PET) [59], Focal Inverse Distance Transform Map for Crowd Localization (FIDTMCL) [41] and Auxiliary Point Guidance Crowd Counting (APGCC) [60].

As illustrated in Fig. 9, representative scenes, including traffic junctions (rows 1-2), snowfields (row 3), dense urban regions (rows 4-5), desert regions (rows 6-7) and other common impervious regions, were selected from 704 validation patches to demonstrate the performance across various scenarios. SCALNet exhibits consistently many missed detections, particularly within the highlighted yellow circles. Similarly, P2PNet produces several false negatives, as exemplified by the cases in rows 2, 5, 8, and 10. PET fails to detect crowds and generates anomalous outputs in densely populated areas (row 9). In contrast, FIDTMCL and APGCC achieve greater localization performance, specifically in complex and high-density regions. The gradual blurring of the signal and lost spatial information on objects, however, limits their further localization performance, as evidenced by missed detections within the yellow circles. With the help of DCPAN and HFGDU, which aim to enhance the feature presentation and recover the spatial information during the upsampling process, CrowdSat-Net outperforms all benchmark methods, achieving the greatest localization performance across all cases.

Quantitative analysis on all 704 validation patches is shown in Table IV. It is clear that CrowdSat-Net achieves state-of-the-art performance with an F1-score of 66.12% and a Precision of 73.23%. Compared to the second-best method FIDTMCL, CrowdSat-Net presents F1-score and Precison gain of 1.72% and 2.42%, respectively, showing its effectiveness in enhancing small-object feature presentation. Notably, CrowdSat-Net outperforms SCALNet by 5.71% in F1-score and 8.41% in Recall, highlighting the effectiveness of its innovative dual-module design. Although APGCC achieves larger Recall, its Precision lags behind CrowdSat-Net by 7.53%, revealing inherent limitations in point-based auxiliary supervision for the precise localization of small objects. Additionally, although PET and P2PNet present great localization performance on images captured by ground surveillance or fixed cameras, they have limitations in detecting small objects on satellite imagery, which are exactly the key challenges that CrowdSat-Net aims to address.

TABLE IV
LOCALIZATION PERFORMANCE OF THE SIX CD METHODS ON THE CROWDSAT DATASET

| Methods | F1-score (%) | Recall (%) | Precision (%) |
|---|---|---|---|
| SCALNet [58] | 60.38 | 51.86 | 72.26 |
| P2PNet [27] | 49.60 | 49.17 | 50.04 |
| PET [59] | 11.36 | 12.54 | 10.41 |
| FIDTMCL [41] | 64.41 | 60.78 | 70.80 |
| APGCC [60] | 64.34 | **63.05** | 65.70 |
| CrowdSat-Net | **66.12** | 60.27 | **73.23** |

The best performance is in bold.

## D. Performance of CrowdSat-Net with Different Crowd Densities

This section examines the performance of CrowdSat-Net systematically across different crowd densities to highlight its adaptive advantages in complex environments. The crowd count ranges were derived from the reference crowd count distribution using quantiles (0.1, 0.3, 0.6, 0.8 0.95). The crowd density levels were then categorized as follows based on these quantile-derived ranges: 1-5 individuals within an image were classified as extremely sparse, 6-20 as sparse, 21-45 as moderate, 46-100 as relatively high, 101-800 as high and 800+ as extremely dense.

As illustrated in Fig. 11, CrowdSat-Net demonstrates consistent robustness in moderate-to-high crowd density scenarios, achieving peak F1-scores of 73.03% in the 46-100 group and 69.60% in the 21-45 group. However, performance degradation was observed in both extremely sparse and extremely dense scenarios. For extremely sparse groups (1-5 individuals), Precision decreases to 35.81%, and in extremely dense crowds (800+), Recall drops to 39.23%. To explain these situations, a visual diagram of these situations is shown in Fig. 11. In extremely sparse scenes, the obvious decrease in Precision is attributed to an increased susceptibility to false positives caused by background clutter, such as image noise and other small objects (e.g., road asphalt, stone pillars and street lamps). Additionally, in extremely dense crowds, the decrease in Recall is due to severe occlusion among individuals, making true positive detection challenging.

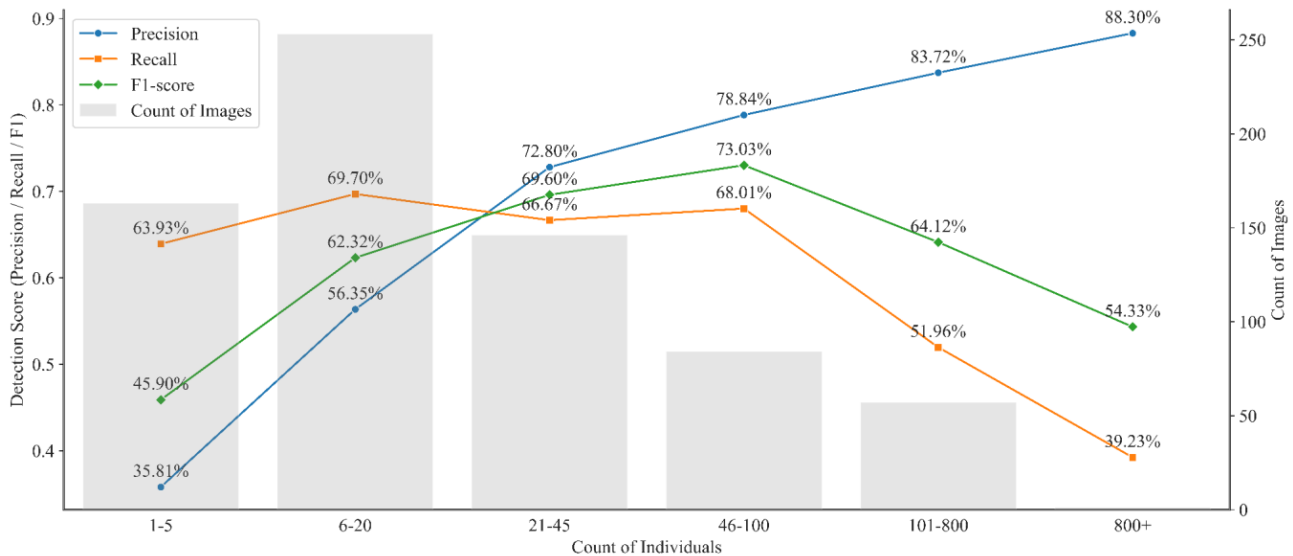

Fig. 9. Performance of CrowdSat-Net on CrowdSat across different crowd densities. The lines show the Precision, Recall and F1-score evaluated on subsets of images grouped by crowd densities, while the grey bars in the background indicate the number of images in each group.

### E. Cross-Regional Generalization of CrowdSat-Net

In this research, the CD dataset with diverse heterogeneity of scenes (e.g., snowy regions, areas of lush vegetation, beaches and desert regions) was assembled to demonstrate fully the generalization capability of CrowdSat-Net. This section tests the cross-regional generalization of CrowdSat-Net in unseen global regions. Six representative global regions, including Red Square in Moscow, Metropolitan Cathedral in Mexico, Phra Nakhon in Bangkok, Djemaa el Fna in Morocco, India Gate in India and National Mall in Washington, were selected for testing.

As displayed in Table V, CrowdSat-Net exhibits notable cross-regional generalization in these scenes. For example, its localization performance in the Metropolitan Cathedral (F1-score: 75.00%; Precision: 82.40%) and the National Mall (F1-score: 73.70%; Recall: 77.74%) is comparable to that in Chinese regions. This is attributed to the diversity of the training dataset and the use of three data augmentation techniques (e.g., horizontal flipping, vertical flipping and CutMix). Additionally, CrowdSat-Net achieves competitive localization performance in the Phra Nakhon, Djemaa el Fna and India Gate.

The visual analysis in Fig. 12 reveals a potential reason for the difference in performance across the six regions: true positives in the six regions were mostly detected in relatively clean backgrounds. For example, in the National Mall, most true positives are located on grass, while in Metropolitan Cathedral, Phra Nakhon and Djemaa el Fna, they are predominantly detected on clean impervious surfaces. In contrast, regions with cluttered backgrounds, such as Red Square and India Gate, exhibited larger false positive and true negative rates. In Red Square, white lane markings on the road obscured individual characteristics, leading to frequent missed detections.

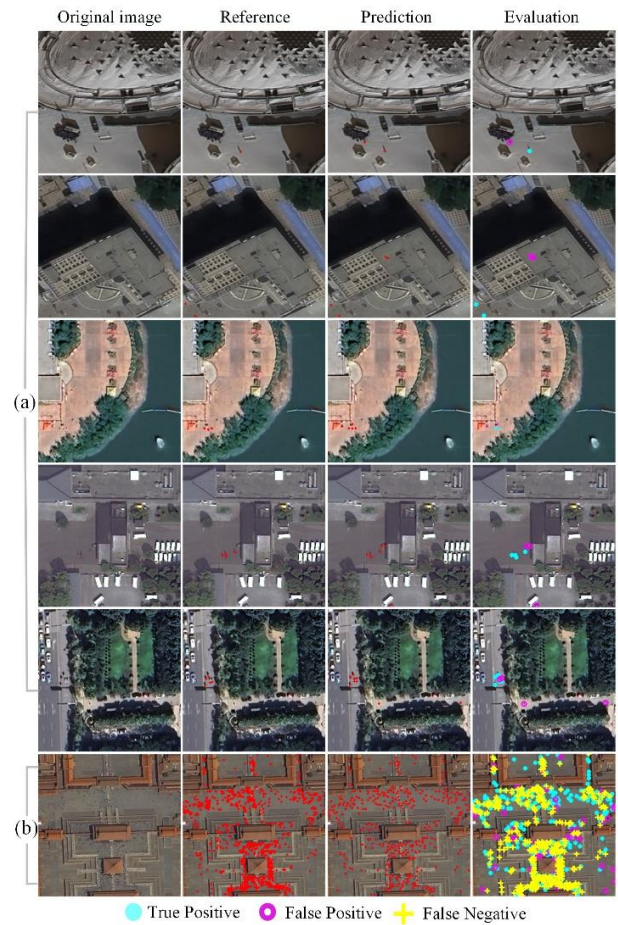

Fig. 10. Examples of the localization performance of CrowdSat-Net in extreme scenarios: (a) Extremely sparse. (b) Extremely dense.

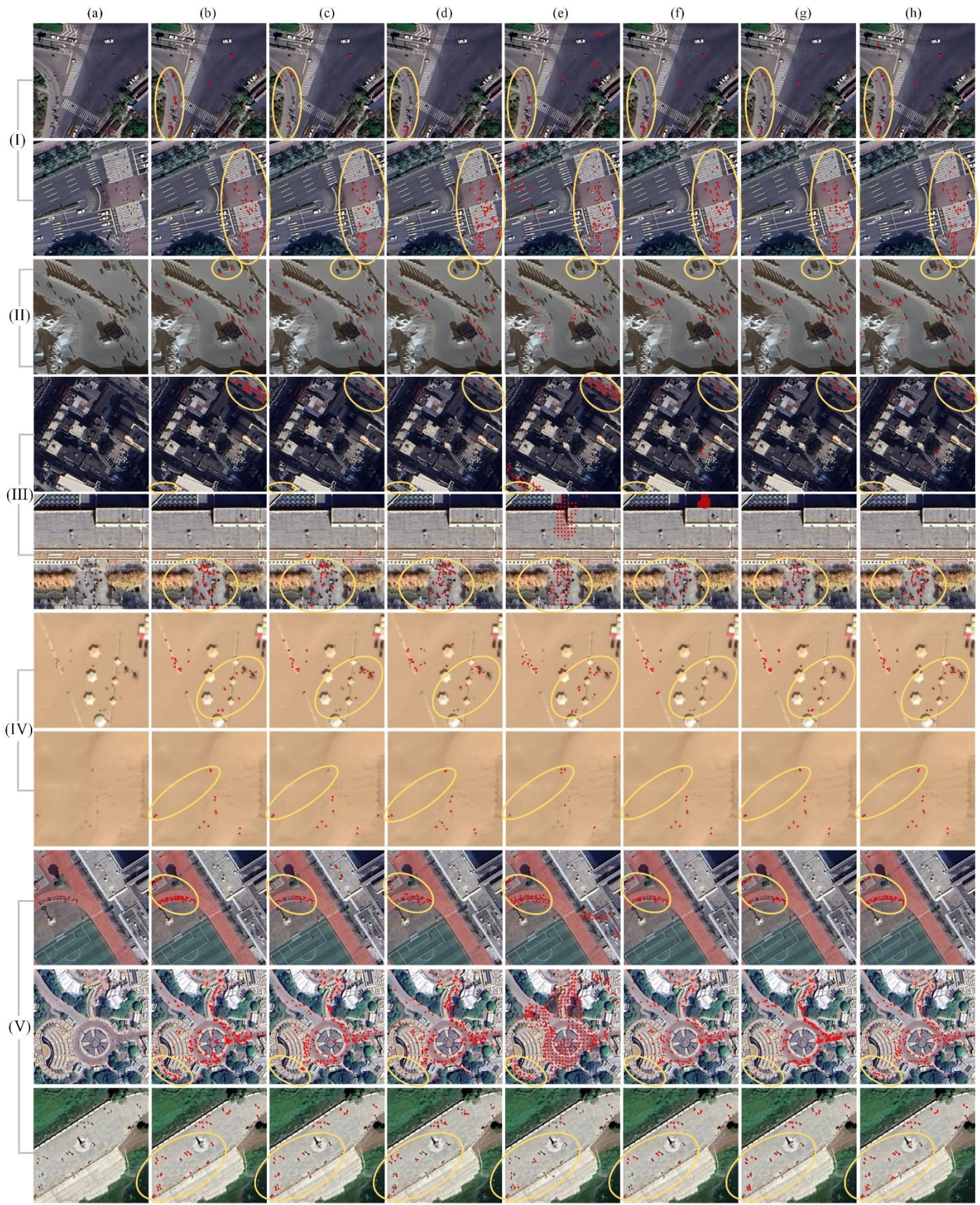

Fig. 11. Visual comparisons between different CD methods for the CrowdSat dataset. (a) Original image. (b) Reference. (c) SCALNet. (d) P2PNet. (e) PET. (f) FIDTMCL. (g) APGCC. (h) CrowdSat-Net. (I) Traffic junctions. (II) Snowfields. (III) Dense urban regions. (IV) Desert regions. (V) Other common impervious regions.

## IV. DISCUSSION

### A. Limitations of the CrowdSat dataset

While CrowdSat demonstrates advantages for CD via VFR satellite imagery, several inherent limitations must be acknowledged to ensure its appropriate application and interpretation. These constraints stem from three fundamental dimensions of satellite observation: spatial resolution, spectral characteristics and temporal sampling.

*Limited Coverage Area.* While CrowdSat was collected in regions with diverse environmental conditions, its coverage range does not exceed China, which may limit the generalizability of models in extremely rare global environments, such as polar regions and dense tropical forests.

*Occlusion Constraints.* The reliance of CrowdSat on optical imagery limits inherently its effectiveness in occluded environments. Physical obstructions such as buildings, dense tree canopies and their shadows can obscure individuals, making crowds only partially detectable or undetectable. This limitation is a fundamental constraint of optical satellite imaging, where line-of-sight visibility is required for accurate detection. Future improvements may explore multi-sensor fusion approaches, such as integrating synthetic aperture radar (SAR) or thermal imaging, to increase performance in occluded environments.

*Minimum Detectable Crowd Size.* At a spatial resolution of 0.3 m, a single pixel accommodates theoretically up to two individuals in extremely high-density scenarios. However, in practice, such cases are exceedingly rare. From the manual labeling of over 120k individual instances, only about 50 cases exhibit densities exceeding this threshold. As current CD models lack the capability to resolve sub-pixel-level individuals, the dataset follows a labeling protocol that constrains annotations to ≤1 individual per pixel for algorithmic compatibility. This limitation may lead to the underestimation of extremely dense gatherings, and future research should explore super-resolution techniques to refine individual-level differentiation.

TABLE V
CROSS-REGIONAL GENERALIZATION OF CROWDSAT-NET IN UNSEEN REGIONS

| Regions | Image Size (pixels) | Date | Source | Spatial Resolution | F1-score (%) | Recall (%) | Precision (%) |
|---|---|---|---|---|---|---|---|
| Red Square | 512 × 512 | Sep. 27, 2023 | Google Earth | 0.3 m | 62.20 | 55.33 | 71.01 |
| Metropolitan Cathedral | 1280 × 1405 | Oct. 11, 2024 | Google Earth | 0.3 m | 75.00 | 68.82 | 82.40 |
| Phra Nakhon | 1402 × 1754 | Jan. 11, 2025 | Google Earth | 0.3 m | 69.17 | 71.03 | 67.41 |
| Djemaa el Fna | 2003 × 1514 | Mar. 14, 2024 | Google Earth | 0.3 m | 69.42 | 71.80 | 67.19 |
| India Gate | 2337 × 677 | Oct. 26, 2024 | Google Earth | 0.3 m | 66.54 | 65.13 | 68.01 |
| National Mall | 2718 × 2079 | Mar. 3, 2024 | Google Earth | 0.3 m | 73.70 | 77.74 | 70.06 |

*Temporal Resolution Constraints.* Although VFR satellites offer relatively frequent revisit times (e.g., BJ3N: every 5 days), the ephemeral nature of crowd activities poses a challenge. Several factors contribute to temporal aliasing in crowd monitoring. Cloud cover and atmospheric conditions can extend significantly the effective revisit interval, sometimes up to 10-15 days, which can delay critical observations. In addition, many satellites, including BJ3N, follow fixed local transit times (such as 11:00 a.m.), which limits their ability to capture nighttime gatherings and events with strong day-night changes, such as evening concerts, religious vigils or mass evacuations after disasters.

### B. Applicability of CrowdSat-Net

In this research, a novel CD method, CrowdSat-Net, designed specifically for satellite imagery, was proposed, coupled with the two key contributions: the DCPAN and HFGDO modules. Through these enhancements, CrowdSat-Net demonstrated reliable localization performance of crowds, and it is expected that CrowdSat-Net and these two modules have the potential for broader applications.

*Generalized Small Object Detection in Satellite Imagery.* While CrowdSat-Net was proposed for large-scale CD, its core design supports generalized small-object detection for satellite imagery. Many non-human objects share similar attributes with human crowds: low pixel occupancy (typically 4-10 pixels per object), irregular spatial distributions and high contextual heterogeneity. Potential applications include 1) wildlife monitoring, such as detecting migratory ungulates, penguin colonies or marine species from VFR satellite imagery; 2) transportation surveillance, including identifying vehicles in road networks, parking lots or logistics hubs; and 3) environmental mapping, such as recognizing vegetation clusters or deforestation patterns in ecological studies. Notably, such non-human objects may have greater detection performance compared to crowds in scenarios where background complexity is reduced. For example, vehicles on uniform asphalt roads or penguin colonies against homogeneous snowy terrain lack the intricate occlusions caused by urban structures (e.g., pillars and poles), enabling clearer feature extraction of object boundaries. Furthermore, some objects, such as orchards or certain naturally occurring species, which tend to be spatially dispersed or organized in regular patterns, have the potential to further reduce spatial interference and increase detection performance.

*Modular Integration with Existing Remote Sensing Frameworks.* While DCPAN and HFGDO were implemented within a stacked Hourglass Network, both modules address fundamental challenges in remote sensing analysis and can be integrated into various backbone architectures. Specifically, DCPAN enhances small-object feature representation, increasing detection robustness in VFR satellite imagery, while HFGDO mitigates spatial detail degradation during upsampling, preserving fine-grained structures. Given these advantages, DCPAN and HFGDO can be embedded into widely used coarse and fine spatial resolution fusion architectures, such as Feature Pyramid Network (FPN) [61], U-net [62] and HRNet [63]. For example, in U-net, each standard skip connection can be embedded with DCPAN and the traditional upsampling method, bilinear interpolation, can be replaced with HFGDO.

### C. Limitations of CrowdSat-Net

While CrowdSat-Net demonstrated notable localization performance for CD, its effectiveness diminishes under extreme crowd densities, exhibiting limitations in both extremely sparse and extremely dense scenarios. As mentioned in Section III-D, in scenarios with extremely sparse crowd distributions, the model exhibits increased sensitivity to small vertical structures, such as thin stone pillars, street lamps and road asphalt. These objects are mis-detected frequently as individuals due to spectral-spatial ambiguity in optical imagery, where individuals and artificial vertical structures exhibit similar spectra. Furthermore, inherent sensor noise exacerbates this ambiguity, reducing the robustness of CrowdSat-Net in such cases. Conversely, in extremely dense crowds, obvious occlusion and signal overlap among individuals result in underestimation errors. As crowd density increases, interpersonal boundaries become less distinguishable, leading to missed detections. This effect is particularly pronounced in high-density gatherings, where individuals exhibit substantial spatial proximity.

### D. Future Research

Based on the limitations identified in CrowdSat and CrowdSat-Net, we propose two key directions for future research:

*Large-Scale and Multi-Source Dataset Expansion.* The CrowdSat dataset primarily covers Chinese regions, which may limit its geographical generalizability, especially for extremely rare global environments. To enhance model robustness, future datasets should incorporate global crowd patterns (by leveraging multi-source optical satellite imagery (~0.3 m spatial resolution) and SAR constellations) and encompass more complex background environments (such as dense urban environments, mixed land cover regions and varied climatic conditions) to improve model adaptability across diverse scenarios.

*Temporal-Aware and Super-Resolution-Based Crowd Detection Frameworks.* To decrease the false detection rates in extremely sparse crowd regions, future models may integrate multi-temporal image sequences with dynamic change detection mechanisms. One possible solution is a dual-branch design, where one part of the model focuses on detecting individuals in an image while the other tracks how they persist over time, reducing false detections due to static, small-sized background objects. On the other hand, to alleviate the individual signal occlusion in extremely dense crowd regions, super-resolution techniques may be employed to enhance image clarity, making it easier to distinguish individuals.

### E. Ethical Considerations

While this research focuses on the use of VFR satellite imagery for constructive and socially beneficial applications, such as urban planning, public safety response, post-disaster response, etc., we acknowledge the broader discussion of the dual-use potential of CD using VFR satellite imagery. Recent research [64] highlights the need to consider the privacy risks associated with fine-resolution remote sensing data carefully. It is important to note that this research utilizes historical satellite imagery, not real-time data. Additionally, while this research can detect the presence of individuals, the individuals are represented as indistinct black or white dot-like shapes in VFR satellite imagery, making it impossible to identify them. Nevertheless, we support future research to develop privacy protection frameworks to ensure the ethical and responsible use of AI-driven geospatial analysis.

## V. CONCLUSION

In this paper, we presented the first study, to the best of our knowledge, that utilized VFR satellite imagery for CD. To achieve this task, we introduced CrowdSat, the first large-scale CD dataset collected via BJ3N and JL4A satellites, as well as the Google Earth platform. CrowdSat comprises over 120k labeled individuals across diverse and heterogeneous environments (e.g., urban areas, snowy regions, lush vegetation, beaches and desert regions). Sourced from 32 provincial-level divisions (except Guizhou Province and Macao) in China, this dataset establishes a foundation for a new, challenging yet meaningful task: detecting large-scale gatherings of individual humans from space.

To address this challenge, we proposed CrowdSat-Net, a novel method specifically designed for CD based on satellite imagery. It incorporates two key contributions: the DCPAN and HFGDO modules. DCPAN improves feature representation for small objects by integrating contextual and local contrast information, while HFGDO mitigates spatial information lost during upsampling. Extensive experiments demonstrated the effectiveness of CrowdSat-Net. The key findings of this paper are summarized as follows.

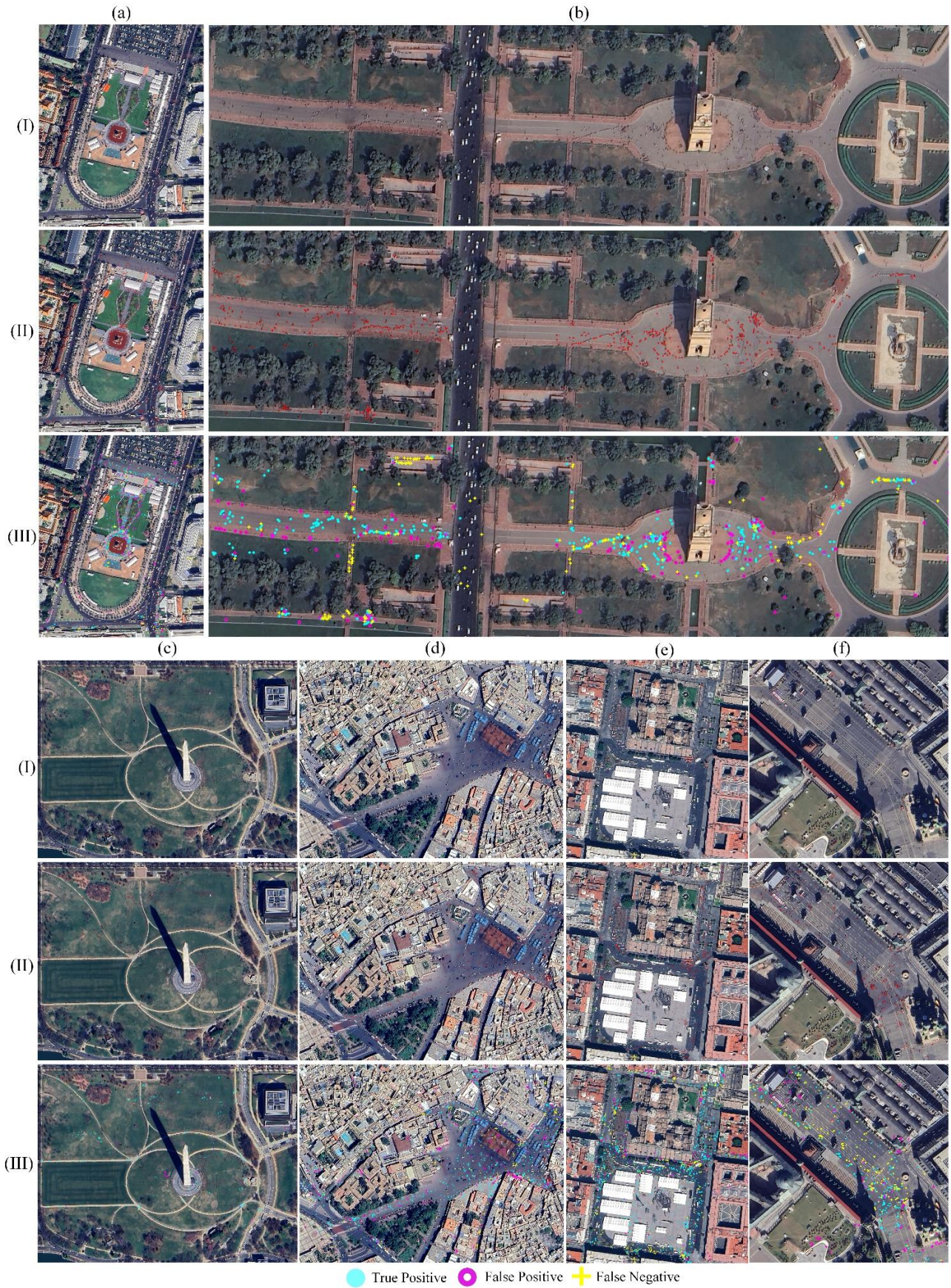

Fig. 12. The visual localization performance of CrowdSat-Net in unseen foreign regions. (a) Phra Nakhon. (b) India Gate. (c) National Mall. (d) Djemaa el Fna. (e) Metropolitan Cathedral. (f) Red Square. (I) Original image. (II) Prediction. (III) Evaluation.

1) VFR satellite imagery overcomes the limitations of small-scale (both spatially and temporally) ground and aerial imagery, offering a promising pathway for large-scale crowd analysis in various applications.

2) CrowdSat-Net outperformed five advanced CD methods (designed for ground or aerial imagery) based on CrowdSat, achieving the largest F1-score of 66.12% and Precision of 73.23%.

3) The DCPAN and HFGDO modules are effective in increasing CD accuracy, increasing the F1-score by 1.70% and Precision by 4.69%.

4) CrowdSat-Net performed reliably in moderate and relatively high crowd density scenarios, with F1-scores of 69.60% and 73.03%, respectively.

5) CrowdSat-Net demonstrated great cross-scene generalization in regions across the globe, with F1-scores of 75.00% and 73.70% in the Metropolitan Cathedral, Mexico and the National Mall, USA, respectively.

Overall, for large-scale CD, VFR satellite imagery offers an appropriate source and CrowdSat-Net provides an effective solution. Future research will focus on constructing larger-scale CD datasets and developing more refined and generalized model architectures.

ACKNOWLEDGMENTS

The authors are grateful to Twenty-First Century Aerospace Technology Co., Ltd. for the BJ3N satellite imagery, Changguang Satellite Technology Co., Ltd. for the JL4A satellite imagery and Google Earth Platform for their satellite data support.